\documentclass[10pt,twocolumn,letterpaper]{article}

\usepackage[pagenumbers]{cvpr} 
\usepackage{url}
\usepackage{graphicx}
\usepackage{booktabs}
\usepackage{graphicx}
\usepackage{multirow}
\usepackage{makecell}
\usepackage{colortbl}
\usepackage[dvipsnames]{xcolor}
\usepackage{amsmath}
\usepackage{amssymb}
\usepackage{dsfont}
\usepackage{mathtools}
\usepackage{enumitem}
\usepackage{microtype}
\DeclareMathOperator*{\argmax}{arg\,max}
\DeclareMathOperator*{\argmin}{arg\,min}

\usepackage{algorithm}
\usepackage{algorithmic}
\definecolor{gray}{gray}{0.5}

\usepackage[pagebackref,breaklinks,colorlinks]{hyperref}

\usepackage[capitalize]{cleveref}
\crefname{section}{Sec.}{Secs.}
\Crefname{section}{Section}{Sections}
\Crefname{table}{Table}{Tables}
\crefname{table}{Tab.}{Tabs.}

\def\cvprPaperID{*****} 
\def\confName{CVPR}
\def\confYear{2022}

\begin{document}

\title{Distill the Best, Ignore the Rest: Improving Dataset Distillation with Loss-Value-Based Pruning}

\author{Brian B. Moser$^{1, 2}$, Federico Raue$^{1}$, Tobias C. Nauen$^{1, 2}$, Stanislav Frolov$^{1, 2}$, Andreas Dengel$^{1, 2}$\\
$^{1}$German Research Center for Artificial Intelligence \\
$^{2}$University of Kaiserslautern-Landau\\
{\tt\small first.last@dfki.de}
}
\maketitle

\begin{abstract}
Dataset distillation has gained significant interest in recent years, yet existing approaches typically distill from the entire dataset, potentially including non-beneficial samples. 
We introduce a novel ``Prune First, Distill After'' framework that systematically prunes datasets via loss-based sampling prior to distillation.
By leveraging pruning before classical distillation techniques and generative priors, we create a representative core-set that leads to enhanced generalization for unseen architectures - a significant challenge of current distillation methods. 
More specifically, our proposed framework significantly boosts distilled quality, achieving up to a 5.2 percentage points accuracy increase even with substantial dataset pruning, \textit{i.e.}, removing 80\% of the original dataset prior to distillation. 
Overall, our experimental results highlight the advantages of our easy-sample prioritization and cross-architecture robustness, paving the way for more effective and high-quality dataset distillation.
\end{abstract}

\section{Introduction}
Large-scale datasets are crucial for training high-quality machine learning models across various applications \cite{rombach2022high,he2016deep}. 
However, the sheer volume of data brings significant computational and storage challenges, making efficient dataset distillation methods highly desirable \cite{cazenavette2022dataset,cui2023scaling}. 
Dataset distillation aims to compress these large datasets into smaller, synthetic subsets while preserving training quality, yet existing techniques often fall short in achieving cross-architecture robustness \cite{moser2024latent}. 
Classifiers generally perform best when their architecture matches the one used during distillation, but performance degrades significantly when trained on other architectures \cite{cazenavette2023generalizing, moser2024latent}. 
This challenge is compounded by the retention of noisy samples, which dilutes the core representational value of distilled data. 
Addressing these issues requires not only compact and representative data but also a selective sampling approach to enhance performance consistency across a range of unseen architectures.

\begin{figure}[t!]
    \begin{center}
        \includegraphics[width=\columnwidth]{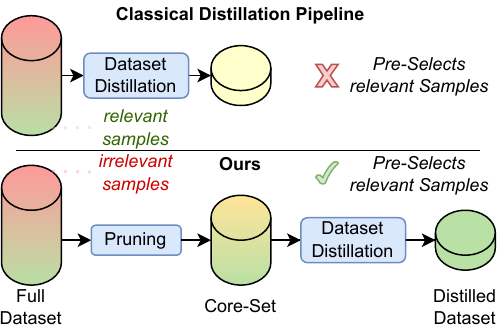}
        \caption{\label{fig:idea}Comparison of classical and our proposed ``Prune First, Distill After'' pipeline: Traditional dataset distillation operates on the full dataset, which includes both relevant and irrelevant samples. Our Prune-Distill approach pre-selects a core-set by pruning loss-value-based irrelevant samples, focusing distillation on the most informative subset, resulting in a more refined distilled dataset.
        }
    \end{center} 
\end{figure}

Inspired by previous work on dataset pruning for various computer vision tasks \cite{moser2024study, moser2022less, ding2023not, coleman2019selection, agarwal2020contextual}, we explore the interplay between dataset pruning and dataset distillation, proposing a novel approach that systematically prunes samples prior to distillation.
As shown in \autoref{fig:idea}, our method enables the creation of compact yet highly representative core-sets, a subset of the original dataset, through targeted pruning that enhance performance and stability across diverse and unseen architectures.
To realize this, we introduce a loss-value-based sampling strategy that leverages a pre-trained classifier model to rank data samples by their ``classification difficulty'', helping to capture the key characteristics of each class.
This analysis combines two sampling strategies: ascending (starting with simpler samples) and descending (starting with complex samples), allowing us to examine their respective effects on distillation quality.
As a result, we find that focusing exclusively on simpler samples yields a significant distillation improvement. 

We build on recent advancements in dataset distillation that incorporate generative priors, exploiting both StyleGAN-XL \cite{cazenavette2023generalizing, sauer2022stylegan} and modified diffusion models \cite{moser2024latent, rombach2022high} to verify our observations. 
Through extensive experiments on subsets of ImageNet, we systematically evaluate the effect of pruning and distillation strategies across diverse architectures.
As a result, we achieve significant performance boosts, sometimes by up to ca. 17\% or 5.2 p.p., on distilled datasets even with substantial pruning factors (\textit{i.e.}, removing 80\% of the original dataset prior to distillation).

Our contributions can be summarized as follows:
\begin{itemize}
    \item A Prune-First, Distill-After Framework: We propose a novel approach that integrates dataset pruning as a pre-processing step for dataset distillation, which improves the quality of the distilled dataset.
    \item Loss-Value-Based Sampling Strategy: We introduce a sampling mechanism based on loss values from a pre-trained classifier, guiding the selection of the most informative samples and enabling class-balanced core-set creation.
    \item Analysis of Pruning Strategies: We compare ascending and descending sampling methods to determine their impact on distillation quality, revealing the benefits of prioritizing simpler samples in the core-set.
    \item Extensive Cross-Architecture Evaluation: We validate our approach on multiple ImageNet subsets and diverse architectures, demonstrating the robustness and flexibility of our pruned and distilled datasets across various model architectures.
\end{itemize}

\section{Background}
This section introduces the key concepts of dataset distillation alongside their fusion with generative priors.
\subsection{Dataset Distillation}
\label{sec:dd}
Let $\mathcal{T} = (X_{real}, Y_{real})$ represent a real dataset with $X_{real} \in \mathbb{R}^{N \times H \times W \times C}$, where $N$ denotes the total number of samples. 
The objective of dataset distillation is to compress $\mathcal{T}$ into a smaller synthetic set $\mathcal{S} = (X_{syn}, Y_{syn})$, with $X_{syn} \in \mathbb{R}^{M \times H \times W \times C}$, where $M = \mathcal{C} \cdot IPC$, $\mathcal{C}$ represents the number of classes, and $IPC$ the images per class. 
Our goal is to achieve $M \ll N$ by optimizing
\begin{equation}\small
\mathcal{S}^* = \arg\min_{\mathcal{S}} \mathcal{L}(\mathcal{S}, \mathcal{T}), \label{eq:def}
\end{equation}
where $\mathcal{L}$ represents the distillation objective, which is defined by the applied dataset distillation method, \textit{e.g.}, Dataset Condensation (DC) \cite{zhao2020dataset}, Distribution Matching (DM) \cite{zhao2023dataset}, and Matching Training Trajectories (MTT) \cite{cazenavette2022dataset}. 

\textbf{Dataset Condensation (DC)} aligns gradients by minimizing the difference between the gradients on the synthetic and real datasets. This is expressed as

\begin{equation}\small
    \mathcal{L}_{DC} (\mathcal{S}, \mathcal{T}) =1- \frac{\nabla_\theta\ell^\mathcal{S}(\theta) \cdot \nabla_\theta\ell^\mathcal{T}(\theta)}{\left\|\nabla_\theta\ell^\mathcal{S}(\theta)\right\|\left\|\nabla_\theta\ell^\mathcal{T}(\theta)\right\|}.
\end{equation}

\textbf{Distribution Matching (DM)} enforces similar feature representations for real and synthetic data by aligning the feature distribution across classes:
\begin{equation}\small
    \mathcal{L}_{DM} (\mathcal{S}, \mathcal{T}) = \sum_{c}\left\|\frac{1}{|\mathcal{T}_c|}\sum_{\mathbf{x}\in \mathcal{T}_c}\psi(\mathbf{x})-\frac{1}{|\mathcal{S}_c|}\sum_{\mathbf{s}\in \mathcal{S}_c}\psi(\mathbf{s})\right\|^2,
\end{equation}
where $\mathcal{T}_c$ and $\mathcal{S}_c$ represent real and synthetic samples for each class $c$.

\textbf{Matching Training Trajectories (MTT)} minimizes the distance between parameter trajectories of networks trained on real and synthetic data. 
Using several model instances, MTT saves the training path of model parameters ${\theta_t^*}_{0}^{T}$ at regular intervals, called expert trajectories. 
For distillation, it initializes a network, $\hat{\theta}_{t+N}$ with $N$ steps, on the synthetic data, tracking its trajectory, and minimizes the distance from the real data trajectory, $\theta^*_{t+M}$ with $M$ steps:
\begin{equation}\small
  \mathcal{L}_{MTT} (\mathcal{S}, \mathcal{T}) = \frac{\|\hat{\theta}_{t+N} - \theta^*_{t+M}\|^2}{\|\theta^*_{t} - \theta^*_{t+M}\|^2}. 
\end{equation}

\subsection{Dataset Distillation with Generative Prior} 
In dataset distillation with a generative prior, a pre-trained generative model is used to synthesize latent codes rather than raw pixel values \cite{cazenavette2023generalizing}.
Incorporating generative priors into dataset distillation offers several advantages, primarily by compressing informative features into a more structured latent space. 
This transformation enables greater flexibility, as latent codes are easier to manipulate and adapt compared to pixel-level data. 
Specifically, let $\mathcal{D}: \mathbb{R}^{M \times h \times w \times d} \rightarrow \mathbb{R}^{M \times H \times W \times C}$ denote the generative model, where $h \cdot w \ll H \cdot W $. 
This transforms the distillation objective as follows:
\begin{equation}\small
\mathcal{Z}^* = \arg\min_{\mathcal{Z}} \mathcal{L}(\mathcal{D}(\mathcal{Z}), \mathcal{T}),
\end{equation}
where $\mathcal{L}$ is the distillation objective defined by the used algorithm (see \autoref{sec:dd}). 
GLaD \cite{cazenavette2023generalizing}, one of the initial methods to leverage generative priors, employs a pre-trained StyleGAN-XL \cite{sauer2022stylegan}.
LD3M \cite{moser2024latent} extends GLaD by replacing the StyleGAN-XL with a modified diffusion model, \textit{i.e.}, Stable Diffusion \cite{rombach2022high}, tailored for dataset distillation.

\subsection{Core-set Selection in Distilled Datasets}
Combining distilled datasets with core-set approaches has already been proposed for small-scale datasets, such as CIFAR100 and Tiny ImageNet.
\textit{He et al.} ~\cite{he2023yoco} suggested applying core-set selection directly to the distilled dataset using two specific rules.
The first rule selects images that are easy to classify based on the logit-based prediction error, while the second rule aims to balance the class distribution using Rademacher complexity.
In addition to this work, \textit{Xu et al.} ~\cite{xu2024distill} introduced an alternative approach that combines core-set selection based on empirical loss before the distillation process with dynamic pruning during the distillation itself.
The main difference between our work is that our model is evaluated on a larger dataset, specifically several subsets of ImageNet, and the combination of core-set selection and distillation process leveraging pre-trained generative models. 
Moreover, we propose to derive core-sets prior to distillation.

\begin{algorithm}[t]
\caption{Loss-Value-Based Sampling}
\label{alg:sampling}
\textbf{Input:} real dataset $\mathcal{T} = (X_{real}, Y_{real})$, pre-trained classifier $\mathcal{M}_\theta$, loss function $\mathcal{L}$, mode $m$ (``easy'' or ``hard''), and pruning ratio $r$.
\begin{algorithmic}[1]
    \STATE \textcolor{gray}{\textit{// initialization}}    
    \STATE \texttt{LossValues} = \texttt{dict}()
    \FOR{ each unique label $\mathbf{y} \in \texttt{SET}(Y_{real})$}
        \STATE \texttt{LossValues}[$\mathbf{y}$] = \texttt{list}()
        \STATE \texttt{DataPairs}[$\mathbf{y}$] = \texttt{list}()
    \ENDFOR
    \STATE 
    \STATE \textcolor{gray}{\textit{// calculate losses}}  
    \FOR{ each sample pair $(\mathbf{x}_i, \mathbf{y}_i) \in \mathcal{T}$}
        \STATE \texttt{LossValues}[$\mathbf{y}_i$].\texttt{append}($\mathcal{L} ( \mathcal{M}_\theta (\mathbf{x}_i), \mathbf{y}_i)$)
        \STATE \texttt{DataPairs}[$\mathbf{y}_i$].\texttt{append}($(\mathbf{x}_i, \mathbf{y}_i)$)
    \ENDFOR
    \STATE
    \STATE \textcolor{gray}{\textit{// derive core-set}}
    \STATE $\pi_r$ = \texttt{list}()
    \FOR{ each unique label $\mathbf{y} \in \texttt{SET}(Y_{real})$}
        \IF {($m$ == ``easy'')}
            \STATE \texttt{indices} = \texttt{argsort} (LossValues[$\mathbf{y}$], ``asc'')
        \ELSE
            \STATE \texttt{indices} = \texttt{argsort} (LossValues[$\mathbf{y}$], ``desc'')
        \ENDIF
        \STATE \texttt{SortedPairs} = \texttt{DataPairs}[\texttt{indices}]
        \STATE $n$ = \texttt{int}($r$ $\cdot$ \texttt{len}(\texttt{SortedPairs}))
        \STATE \texttt{PrunedPairs} = \texttt{SortedPairs}[:n]
        \STATE $\pi_r$ = \texttt{cat} ($\pi_r$, \texttt{PrunedPairs})
    \ENDFOR
    \STATE
    \STATE \textbf{Return:} pruned dataset $\pi_r$
\end{algorithmic}
\end{algorithm}

\section{Methodology}
\begin{figure}[t!]
    \begin{center}
        \includegraphics[width=\columnwidth]{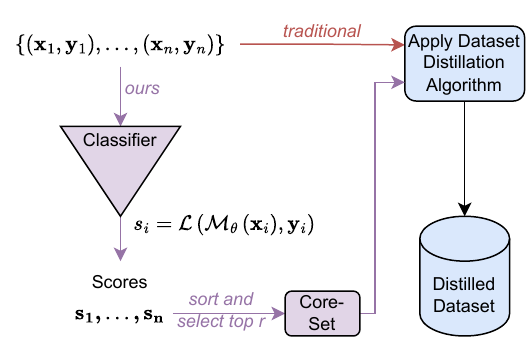}
        \caption{\label{fig:method}Comparison of traditional dataset distillation with our proposed ``Prune First, Distill After'' pipeline. Rather than distilling the full dataset, we first use a classifier to rank and select the top-r samples to create a core-set, which is then distilled to produce a more expressive and architecture-robust distilled dataset.
        }
    \end{center} 
\end{figure}

The overarching goal of dataset distillation is to reduce the size of a dataset by learning strong and representative synthetic samples.
Yet, finding the most representative and informative features also involves the complementary task of removing noisy and poor information.
Thus, we propose dataset pruning as a crucial pre-processing step for effective dataset distillation. 
In this section, we introduce the concept of core-sets and present our method for constructing optimized core-sets for dataset distillation, as shown in \autoref{fig:method} and outlined in \autoref{alg:sampling}.

\subsection{Core-Sets}
Let $\mathcal{T} = (X_{real}, Y_{real})$ represent a dataset of size $\text{N}_\mathcal{T}$, where $\mathbf{x}_i \in X_{real}$ denotes the $i$-th image sample and $\mathbf{y}_i \in Y_{real}$ its label. 
A core-set $\pi_r \subset \mathcal{T}$ is defined as a subset of size $\text{N}_{\pi_r} \approx r \cdot \text{N}_\mathcal{T}$, with $r \in (0, 1)$ determining its relative size.

The goal is to construct a core-set that improves the quality of distilled datasets by focusing on key samples, achieving the specified subset size through a sampling strategy that satisfies the condition
\begin{equation}
    \text{N}_{\pi_r} = \sum_{ \left(\mathbf{x}_i, \mathbf{y}_i \right) \in \mathcal{T}} \mathds{1}_{\pi_r} \left( \mathbf{x}_i \right) \approx r \cdot \text{N}_\mathcal{T},
    \label{eq:sampling}
\end{equation}
where $\mathds{1}_{\pi_r}: \mathcal{T} \rightarrow \{ 0, 1\}$ serves as an indicator function, marking membership of elements in the core-set $\pi_r$ within the larger dataset $\mathcal{T}$.
In other words, the core-set $\pi_r$ acts between the original dataset and the synthetic dataset, representing only a fraction of the original dataset to the dataset distillation methods.
The concrete realization of the core-set depends on the sampling mechanism that defines the indicator function $\mathds{1}_{\pi_r}$.

\subsection{Loss-Value-based Sampling}
In image classification, one tries to approximate $\mathbf{y}_i \approx \mathcal{M}_\theta \left( \mathbf{x}_i \right)$, where $\mathcal{M}_\theta$ is a classifier with parameters $\theta$ (\textit{i.e.}, CNN models like AlexNet, VGG, ResNet, etc.).
Traditionally, we try to optimize $\mathcal{M}_\theta$ by minimizing the loss $\mathcal{L} (\mathcal{M}_\theta \left( \mathbf{x}_i \right), \mathbf{y}_i)$ between the predicted labels and the actual labels. 
However, one can also use the loss of a trained model $\mathcal{M}_\theta$ as a metric for ``classification difficulty''.
Building upon this idea and given $r \in \left( 0, 1\right)$, we can derive a core-set with
\begin{equation}
    \label{eq:easy}
    \pi^{\text{easy}}_r = \argmin_{\substack{\mathcal{T}' \subset \mathcal{T}, \\ \mid \mathcal{T}'\mid \approx r \cdot \text{N}_\mathcal{T}}} \sum_{\left(\mathbf{x}_i, \mathbf{y}_i \right) \in \mathcal{T}'} \mathcal{L} \left( \mathcal{M}_\theta \left( \mathbf{x}_i \right), \mathbf{y}_i\right),
\end{equation}
where samples with high loss values are removed first.
Vice versa, we can define the opposite extreme case by removing the lowest loss values first, with
\begin{equation}
    \label{eq:hard}
    \pi^{\text{hard}}_r = \argmax_{\substack{\mathcal{T}' \subset \mathcal{T}, \\ \mid \mathcal{T}'\mid \approx r \cdot \text{N}_\mathcal{T}}} \sum_{\left(\mathbf{x}_i, \mathbf{y}_i \right) \in \mathcal{T}'} \mathcal{L} \left( \mathcal{M}_\theta \left( \mathbf{x}_i\right), \mathbf{y}_i\right).
\end{equation}

\begin{table*}[t!]
\centering
\resizebox{\textwidth}{!}{%
\begin{tabular}{lccccccccccc}
Distillation Space & Algorithm & ImNet-A & ImNet-B & ImNet-C & ImNet-D & ImNet-E & ImNette & ImWoof & ImNet-Birds & ImNet-Fruits & ImNet-Cats         \\\midrule
 & MTT & 33.4{\color{gray}$\pm$1.5} & 34.0{\color{gray}$\pm$3.4} & 31.4{\color{gray}$\pm$3.4} & 27.7{\color{gray}$\pm$2.7} & 24.9{\color{gray}$\pm$1.8} & 24.1{\color{gray}$\pm$1.8} & 16.0{\color{gray}$\pm$1.2} & 25.5{\color{gray}$\pm$3.0} & 18.3{\color{gray}$\pm$2.3} & 18.7{\color{gray}$\pm$1.5} \\
pixel space & DC & 38.7{\color{gray}$\pm$4.2} & 38.7{\color{gray}$\pm$1.0} & 33.3{\color{gray}$\pm$1.9} & 26.4{\color{gray}$\pm$1.1} & 27.4{\color{gray}$\pm$0.9} & 28.2{\color{gray}$\pm$1.4} & 17.4{\color{gray}$\pm$1.2} & 28.5{\color{gray}$\pm$1.4} & 20.4{\color{gray}$\pm$1.5} & 19.8{\color{gray}$\pm$0.9} \\
 & DM & 27.2{\color{gray}$\pm$1.2} & 24.4{\color{gray}$\pm$1.1} & 23.0{\color{gray}$\pm$1.4} & 18.4{\color{gray}$\pm$0.7} & 17.7{\color{gray}$\pm$0.9} & 20.6{\color{gray}$\pm$0.7} & 14.5{\color{gray}$\pm$0.9} & 17.8{\color{gray}$\pm$0.8} & 14.5{\color{gray}$\pm$1.1} & 14.0{\color{gray}$\pm$1.1} \\
 \midrule \midrule
 
 & MTT & 39.9{\color{gray}$\pm$1.2} & 39.4{\color{gray}$\pm$1.3} & 34.9{\color{gray}$\pm$1.1} & 30.4{\color{gray}$\pm$1.5} & 29.0{\color{gray}$\pm$1.1} & 30.4{\color{gray}$\pm$1.5} & 17.1{\color{gray}$\pm$1.1} & 28.2{\color{gray}$\pm$1.1} & 21.1{\color{gray}$\pm$1.2} & 19.6{\color{gray}$\pm$1.2} \\
GLaD & DC & 41.8{\color{gray}$\pm$1.7} & 42.1{\color{gray}$\pm$1.2} & 35.8{\color{gray}$\pm$1.4} & 28.0{\color{gray}$\pm$0.8} & 29.3{\color{gray}$\pm$1.3} & 31.0{\color{gray}$\pm$1.6} & 17.8{\color{gray}$\pm$1.1} & 29.1{\color{gray}$\pm$1.0} & 22.3{\color{gray}$\pm$1.6} & 21.2{\color{gray}$\pm$1.4} \\
 & DM & 31.6{\color{gray}$\pm$1.4} & 31.3{\color{gray}$\pm$3.9} & 26.9{\color{gray}$\pm$1.2} & 21.5{\color{gray}$\pm$1.0} & 20.4{\color{gray}$\pm$0.8} & 21.9{\color{gray}$\pm$1.1} & 15.2{\color{gray}$\pm$0.9} & 18.2{\color{gray}$\pm$1.0} & 20.4{\color{gray}$\pm$1.6} & 16.1{\color{gray}$\pm$0.7} \\ \midrule

& MTT & 40.4{\color{gray}$\pm$1.6} & 42.4{\color{gray}$\pm$1.3} & 35.4{\color{gray}$\pm$1.1} & 30.5{\color{gray}$\pm$1.3} & 30.0{\color{gray}$\pm$1.1} & 30.5{\color{gray}$\pm$1.0} & 20.3{\color{gray}$\pm$1.3} & 27.7{\color{gray}$\pm$1.6} & 21.8{\color{gray}$\pm$1.1} & 21.2{\color{gray}$\pm$0.8} \\
& & \color{ForestGreen}+0.5{\color{gray}+0.4} & \color{ForestGreen}+3.0{\color{gray}+0.0} & \color{ForestGreen}+0.6{\color{gray}+0.0} & \color{ForestGreen}+0.1{\color{gray}-0.2} & \color{ForestGreen}+1.0{\color{gray}+0.0} & \color{ForestGreen}+0.1{\color{gray}-0.5} & \color{ForestGreen}+3.2{\color{gray}+0.2} & \color{BrickRed}-0.5{\color{gray}-0.5} & \color{ForestGreen}+0.7{\color{gray}-0.1} & \color{ForestGreen}+0.6{\color{gray}-0.4} \\

\textbf{GLaD} & DC & 42.8{\color{gray}$\pm$1.9} & 44.6{\color{gray}$\pm$1.0} & 37.6{\color{gray}$\pm$2.0} & 29.5{\color{gray}$\pm$1.7} & 32.8{\color{gray}$\pm$1.0} & 33.6{\color{gray}$\pm$1.0} & 19.3{\color{gray}$\pm$1.2} & 30.9{\color{gray}$\pm$0.9} & 23.3{\color{gray}$\pm$1.1} & 22.4{\color{gray}$\pm$0.8} \\
(pruned)& & \color{ForestGreen}+1.0{\color{gray}+0.5} & \color{ForestGreen}+2.5{\color{gray}-0.2} & \color{ForestGreen}+2.2{\color{gray}+0.9} & \color{ForestGreen}+1.5{\color{gray}+0.9} & \color{ForestGreen}+3.5{\color{gray}-0.3} & \color{ForestGreen}+2.6{\color{gray}-0.6} & \color{ForestGreen}+1.5{\color{gray}-0.1} & \color{ForestGreen}+1.8{\color{gray}-0.1} & \color{ForestGreen}+1.0{\color{gray}-0.5} & \color{ForestGreen}+1.2{\color{gray}-0.6} \\

 & DM & 36.8{\color{gray}$\pm$1.6} & 36.5{\color{gray}$\pm$1.6} & 30.8{\color{gray}$\pm$1.2} & 23.1{\color{gray}$\pm$1.4} & 22.8{\color{gray}$\pm$2.0} & 24.9{\color{gray}$\pm$0.8} & 16.1{\color{gray}$\pm$0.9} & 21.4{\color{gray}$\pm$0.9} & 22.7{\color{gray}$\pm$1.1} & 16.3{\color{gray}$\pm$1.0} \\
 & & \color{ForestGreen}+5.2{\color{gray}+0.2} & \color{ForestGreen}+5.2{\color{gray}-2.3} & \color{ForestGreen}+3.9{\color{gray}+0.0} & \color{ForestGreen}+1.6{\color{gray}+0.4} & \color{ForestGreen}+2.4{\color{gray}+1.2} & \color{ForestGreen}+3.0{\color{gray}-0.3} & \color{ForestGreen}+0.9{\color{gray}+0.0} & \color{ForestGreen}+3.2{\color{gray}-0.1} & \color{ForestGreen}+2.3{\color{gray}-0.5} & \color{ForestGreen}+0.2{\color{gray}+0.3} \\
 \midrule \midrule
 
 & MTT & 40.9{\color{gray}$\pm$1.1} & 41.6{\color{gray}$\pm$1.7} & 34.1{\color{gray}$\pm$1.7} & 31.5{\color{gray}$\pm$1.2} & 30.1{\color{gray}$\pm$1.3} & 32.0{\color{gray}$\pm$1.3} & 19.9{\color{gray}$\pm$1.2} & 30.4{\color{gray}$\pm$1.5} & 21.4{\color{gray}$\pm$1.1} & 22.1{\color{gray}$\pm$1.0} \\
LD3M & DC & 42.3{\color{gray}$\pm$1.3} & 42.0{\color{gray}$\pm$1.1} & 37.1{\color{gray}$\pm$1.8} & 29.7{\color{gray}$\pm$1.3} & 31.4{\color{gray}$\pm$1.1} & 32.9{\color{gray}$\pm$1.2} & 18.9{\color{gray}$\pm$0.6} & 30.2{\color{gray}$\pm$1.4} & 22.6{\color{gray}$\pm$1.3} & 21.7{\color{gray}$\pm$0.8} \\
 & DM & 35.8{\color{gray}$\pm$1.1} & 34.1{\color{gray}$\pm$1.0} & 30.3{\color{gray}$\pm$1.2} & 24.7{\color{gray}$\pm$1.0} & 24.5{\color{gray}$\pm$0.9} & 26.8{\color{gray}$\pm$1.7} & 18.1{\color{gray}$\pm$0.7} & 23.0{\color{gray}$\pm$1.8} & 24.5{\color{gray}$\pm$1.9} & 17.0{\color{gray}$\pm$1.1} \\ \midrule

& MTT & 41.2{\color{gray}$\pm$1.5} & 42.4{\color{gray}$\pm$1.2} & 36.0{\color{gray}$\pm$1.4} & 31.7{\color{gray}$\pm$1.4} & 29.9{\color{gray}$\pm$0.9} & 33.8{\color{gray}$\pm$1.6} & 19.9{\color{gray}$\pm$1.2} & 29.5{\color{gray}$\pm$1.1} & 23.3{\color{gray}$\pm$1.2} & 22.3{\color{gray}$\pm$0.9} \\
& & \color{ForestGreen}+0.3{\color{gray}+0.5} & \color{ForestGreen}+0.8{\color{gray}-0.5} & \color{ForestGreen}+1.9{\color{gray}-0.3} & \color{ForestGreen}0.2{\color{gray}+0.2} & \color{BrickRed}-0.2{\color{gray}-0.4} & \color{ForestGreen}+1.8{\color{gray}+0.3} & +0.0{\color{gray}+0.0} & \color{BrickRed}-0.9{\color{gray}-0.4} & \color{ForestGreen}+1.9{\color{gray}+0.1} & \color{ForestGreen}+0.2{\color{gray}-0.1} \\

\textbf{LD3M} & DC & 42.8{\color{gray}$\pm$1.8} & 44.0{\color{gray}$\pm$1.4} & 37.8{\color{gray}$\pm$1.1} & 29.7{\color{gray}$\pm$0.9} & 33.1{\color{gray}$\pm$1.3} & 34.4{\color{gray}$\pm$1.3} & 19.4{\color{gray}$\pm$0.9} & 31.6{\color{gray}$\pm$1.3} & 24.0{\color{gray}$\pm$1.0} & 22.5{\color{gray}$\pm$1.1} \\
(pruned)&  & \color{ForestGreen}+0.5{\color{gray}+0.5} & \color{ForestGreen}+2.0{\color{gray}+0.3} & \color{ForestGreen}+0.7{\color{gray}-0.7} & +0.0{\color{gray}-0.4} & \color{ForestGreen}+1.7{\color{gray}+0.2} & \color{ForestGreen}+1.5{\color{gray}+0.1} & \color{ForestGreen}+0.5{\color{gray}+0.3} & \color{ForestGreen}+1.4{\color{gray}-0.1} & \color{ForestGreen}+1.4{\color{gray}-0.3} & \color{ForestGreen}+0.8{\color{gray}+0.3} \\

 & DM & 38.8{\color{gray}$\pm$1.2} & 34.9{\color{gray}$\pm$1.4} & 29.7{\color{gray}$\pm$1.6} & 26.0{\color{gray}$\pm$1.6} & 27.2{\color{gray}$\pm$0.9} & 29.6{\color{gray}$\pm$1.3} & 18.4{\color{gray}$\pm$1.0} & 24.5{\color{gray}$\pm$1.3} & 24.5{\color{gray}$\pm$1.4} & 17.3{\color{gray}$\pm$1.2} \\
 & & \color{ForestGreen}+3.0{\color{gray}+0.1} & \color{ForestGreen} +0.8{\color{gray}+0.4} & \color{BrickRed}-0.5{\color{gray}+0.4} & \color{ForestGreen}+1.3{\color{gray}+0.6} & \color{ForestGreen}+2.7{\color{gray}+0.0} & \color{ForestGreen}+2.8{\color{gray}-0.4} & \color{ForestGreen}+0.3{\color{gray}+0.3} & \color{ForestGreen}+1.5{\color{gray}-0.5} & +0.0{\color{gray}-0.5} & \color{ForestGreen}+0.3{\color{gray}+0.1} \\
\end{tabular}
}
\caption{
Cross-architecture performance on ImageNet (128×128) subsets with 1 Image Per Class (IPC). Distillation methods (MTT, DC, DM) were evaluated using AlexNet, VGG11, ResNet18, and ViT as unseen architectures, with performance averaged over actual validation sets. Generative priors, GLaD and LD3M, were tested on both full datasets and pruned core-sets, showing relative changes in performance (green for improvements, red for declines). Notably, pruning almost always led to performance improvements across various methods and subsets, highlighting the benefit of core-set selection in dataset distillation.
}
\label{tab:main_results}
\end{table*}

\begin{figure*}[t!]
    \begin{center}
        \includegraphics[width=\textwidth]{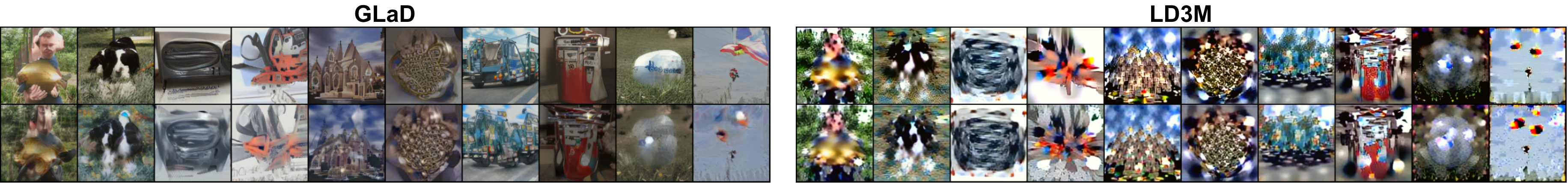}
        \caption{\label{fig:pruned_vis}IPC=1 class images from ImageNette (tench, English springer, cassette player, chain saw, church, French horn, garbage truck, gas pump, golf ball, parachute) using MTT with GLaD \textbf{(left)} and LD3M \textbf{(right)}. The top row shows images distilled from the full dataset, while the bottom row displays distilled images after pruning ($r=0.6$).
        }
    \end{center} 
\end{figure*}

In summary, for $\pi_r^{\text{easy}}$, low-loss-value samples - often representing ``easier'' instances - are prioritized. 
As such, they can help the model establish core class characteristics without excessive complexity, ensuring a distilled dataset that captures essential features.
On the other hand, for $\pi_r^{\text{hard}}$, high-loss-value samples - often complex, high-variation instances - are included first.
These samples, rich in nuance and feature diversity, are useful for learning fine distinctions within a class. 
However, introducing such variability early risks adding unnecessary complexity, which may hinder effective distillation by overwhelming the core representation.

\section{Experimental Setup}
\label{sec:setup}
We follow \textit{Cazenavette et al.} \cite{cazenavette2023generalizing} and evaluate the cross-architecture performance with generative priors (GLaD and LD3M), for IPC=1 (MTT, DC, DM) and IPC=10 (DC, DM) with image size $128\times128$ as well as an evaluation with DC and image size $256\times256$ for IPC=1.
In all experiments, we maintain consistent hyperparameters to guarantee a fair comparison. Our code can be found on GitHub\footnote{\url{https://github.com/Brian-Moser/prune_and_distill}}.

\subsection{Datasets}
We assess classifier accuracy on synthetic images from various 10-class subsets of ImageNet-1k \cite{deng2009imagenet}. 
The subsets include ImageNet-Birds, ImageNet-Fruits, and ImageNet-Cats, as defined in prior work \cite{cazenavette2022dataset}, as well as two widely used subsets, ImageNette and ImageWoof \cite{howard2019smaller}. Additionally, we use subsets based on ResNet-50 performance on ImageNet \cite{cazenavette2023generalizing}: ImageNet-A contains the top 10 classes, followed by ImageNet-B, ImageNet-C, ImageNet-D, and ImageNet-E. 

\subsection{Evaluation Protocol}
We begin by distilling synthetic datasets using the chosen algorithms (\textit{i.e.}, DC, DM, and MTT), followed by assessing their quality across previously unseen network architectures. 
To evaluate each synthetic dataset with a particular architecture, a new network is trained from scratch on the distilled dataset and then tested on real images in the test set. 
This process is repeated five times, and we report the mean test accuracy with a confidence interval of plus or minus one standard deviation.

\subsection{Network Architectures}
Following \textit{Cazenavette et al.} \cite{cazenavette2023generalizing} and previous work in dataset distillation \cite{nguyen2021dataset, cui2023scaling, nguyen2020dataset, moser2024latent}, we use ConvNet to distill the $128\times128$ and $256\times256$ datasets, respectively \cite{gidaris2018dynamic}. 
For evaluating unseen architectures, we employ AlexNet \cite{krizhevsky2012imagenet}, VGG-11 \cite{simonyan2014very}, ResNet-18 \cite{he2016deep}, and a Vision Transformer \cite{dosovitskiy2020image}.

\subsection{Generative Priors}
We incorporate two types of generative priors: GLaD \cite{cazenavette2023generalizing}, which leverages a StyleGAN-XL \cite{sauer2022stylegan} model, and LD3M \cite{moser2024latent}, which uses a modified Stable Diffusion \cite{rombach2022high} model adapted specifically for dataset distillation. 

\subsection{Pruning Details}
We applied a pre-trained ResNet-18 model \cite{he2016deep} with cross-entropy loss as the scoring mechanism for our proposed loss-value-based sampling. 

\section{Results}
This section evaluates the impact of our pruning approach across several ImageNet subsets and distillation methods, where we used $\pi_r^{\text{easy}}$ ($r=0.2$ for DM and $r=0.6$ for DC and MTT). 
Next, we present our analysis of $\pi_r^{\text{easy}}$ and $\pi_r^{\text{hard}}$, from which we derive the optimal settings and values for $r$.

\subsection{Distillation Results}
This section demonstrates the impact of our pruning approach on the performance of dataset distillation across multiple ImageNet subsets and architectures. 
The pruning settings (60\% for DC, MTT, and 20\% for DM) were derived from the analysis provided in the next \autoref{sec:pa}.
Overall, the proposed loss-based pruning consistently improves performance across various distillation algorithms.

\textbf{IPC=1, 128$\times$128.} As shown in \autoref{tab:main_results}, applying pruning yields significant accuracy gains across nearly all subsets, with particularly strong improvements for the DC and DM algorithms for IPC=1. 
Specifically, pruning prior to DM with GLaD improves accuracy by +5.2 percentage points on the ImageNet-A and B subsets and by +3.9 percentage points on ImageNet-C. 
Similarly, DM with LD3M sees notable boosts, particularly on ImageNet-A (+3.0 points) and ImageNette (+2.8 points). 
Overall, GLaD demonstrates more pronounced accuracy improvements than LD3M, emphasizing the effectiveness of pruning and generalizability across diverse datasets and unseen architectures.
With pruning, it is even possible to boost GLaD beyond the performance of LD3M; for example, on the ImageWoof subset with MTT, pruned GLaD achieves a slight improvement of +0.4 percentage points over pruned LD3M.

\textbf{Visuals for IPC=1, 128$\times$128.} In \autoref{fig:pruned_vis}, the distilled class images are shown for GLaD and LD3M, respectively (ImageNette, MTT).
The visualizations highlight the changes in distilled images when applying pruning to select a core set. 
In both cases, GLaD and LD3M, the images distilled from the core-set (bottom row) exhibit a more refined depiction of class-specific features compared to the full dataset (top row).
Overall, the changes in distilled images are more pronounced in GLaD, which aligns with the observation of greater accuracy improvements listed in \autoref{tab:main_results}.

\begin{table*}[t!]
\centering
\resizebox{\textwidth}{!}{%
\begin{tabular}{lccccccc}
Distillation Space & Algorithm & All & ImNet-A & ImNet-B & ImNet-C & ImNet-D & ImNet-E \\ \midrule
 & DC & 42.3{\color{gray}$\pm$3.5} & 52.3{\color{gray}$\pm$0.7} & 45.1{\color{gray}$\pm$8.3} & 40.1{\color{gray}$\pm$7.6} & 36.1{\color{gray}$\pm$0.4} & 38.1{\color{gray}$\pm$0.4} \\
\multirow{-2}{*}{pixel space} & DM & 44.4{\color{gray}$\pm$0.5} & 52.6{\color{gray}$\pm$0.4} & 50.6{\color{gray}$\pm$0.5} & 47.5{\color{gray}$\pm$0.7} & 35.4{\color{gray}$\pm$0.4} & 36.0{\color{gray}$\pm$0.5} \\ \midrule\midrule
 & DC & 45.9{\color{gray}$\pm$1.0} & 53.1{\color{gray}$\pm$1.4} & 50.1{\color{gray}$\pm$0.6} & 48.9{\color{gray}$\pm$1.1} & 38.9{\color{gray}$\pm$1.0} & 38.4{\color{gray}$\pm$0.7} \\
 
\multirow{-2}{*}{GLaD} & DM & 45.8{\color{gray}$\pm$0.6} & 52.8{\color{gray}$\pm$1.0} & 51.3{\color{gray}$\pm$0.6} & 49.7{\color{gray}$\pm$0.4} & 36.4{\color{gray}$\pm$0.4} & 38.6{\color{gray}$\pm$0.7} \\ \midrule
 & DC & 47.2{\color{gray}$\pm$1.0} & 54.7{\color{gray}$\pm$1.1} & 52.3{\color{gray}$\pm$0.8} & 48.5{\color{gray}$\pm$1.3} & 39.7{\color{gray}$\pm$0.8} & 40.6{\color{gray}$\pm$1.0} \\
 
\textbf{GLaD} &  & \color{ForestGreen}+1.3{\color{gray}+0.0} & \color{ForestGreen}+1.6{\color{gray}-0.3} & \color{ForestGreen}+2.2{\color{gray}+0.2} & \color{BrickRed}-0.4{\color{gray}+0.2} & \color{ForestGreen}+0.8{\color{gray}-0.2} & \color{ForestGreen}+2.2{\color{gray}+0.3} \\ 

(pruned) & DM & 46.8{\color{gray}$\pm$1.2} & 55.3{\color{gray}$\pm$1.3} & 51.7{\color{gray}$\pm$1.0} & 49.1{\color{gray}$\pm$0.8} & 38.5{\color{gray}$\pm$1.1} & 39.6{\color{gray}$\pm$1.6} \\
&  & \color{ForestGreen}+1.0{\color{gray}+0.6} & \color{ForestGreen}+2.5{\color{gray}+0.3} & \color{ForestGreen}+0.4{\color{gray}+0.4} & \color{BrickRed}-0.6{\color{gray}+0.4} & \color{ForestGreen}+1.9{\color{gray}+0.7} & \color{ForestGreen}+1.0{\color{gray}+0.9} \\ \midrule\midrule

 & DC & 47.1{\color{gray}$\pm$1.2} & 55.2{\color{gray}$\pm$1.0} & 51.8{\color{gray}$\pm$1.4} & 49.9{\color{gray}$\pm$1.3} & 39.5{\color{gray}$\pm$1.0} & 39.0{\color{gray}$\pm$1.3} \\
\multirow{-2}{*}{LD3M} & DM & 47.3{\color{gray}$\pm$2.1} & 57.0{\color{gray}$\pm$1.3} & 52.3{\color{gray}$\pm$1.1} & 48.2{\color{gray}$\pm$4.9} & 39.5{\color{gray}$\pm$1.5} & 39.4{\color{gray}$\pm$1.8} \\ \midrule

 & DC & 47.8{\color{gray}$\pm$1.1} & 55.4{\color{gray}$\pm$1.3} & 53.5{\color{gray}$\pm$1.0} & 49.5{\color{gray}$\pm$1.0} & 40.1{\color{gray}$\pm$1.3} & 40.6{\color{gray}$\pm$0.9} \\
\textbf{LD3M} &  & \color{ForestGreen}+0.7{\color{gray}-0.1} & \color{ForestGreen}+0.2{\color{gray}+0.3} & \color{ForestGreen}+1.7{\color{gray}-0.4} & \color{BrickRed}-0.4{\color{gray}-0.3} & \color{ForestGreen}+0.6{\color{gray}+0.3} & \color{ForestGreen}+1.6{\color{gray}-0.4} \\ 

(pruned) & DM & 48.7{\color{gray}$\pm$1.3} & 58.0{\color{gray}$\pm$1.1} & 55.6{\color{gray}$\pm$1.2} & 48.3{\color{gray}$\pm$1.5} & 39.7{\color{gray}$\pm$1.2} & 41.8{\color{gray}$\pm$1.3} \\
&  & \color{ForestGreen}+1.4{\color{gray}-0.8} & \color{ForestGreen}+1.0{\color{gray}-0.2} & \color{ForestGreen}+3.3{\color{gray}+0.1} & \color{ForestGreen}+0.1{\color{gray}-3.4} & \color{ForestGreen}+0.2{\color{gray}-0.3} & \color{ForestGreen}+2.4{\color{gray}-0.5} 
\end{tabular}
}
\caption{Cross-architecture performance with IPC=10 on the subsets ImageNet A to E.
}
\label{tab:10ipc}
\end{table*}

\begin{figure*}[t!]
    \begin{center}
        \includegraphics[width=.84\textwidth]{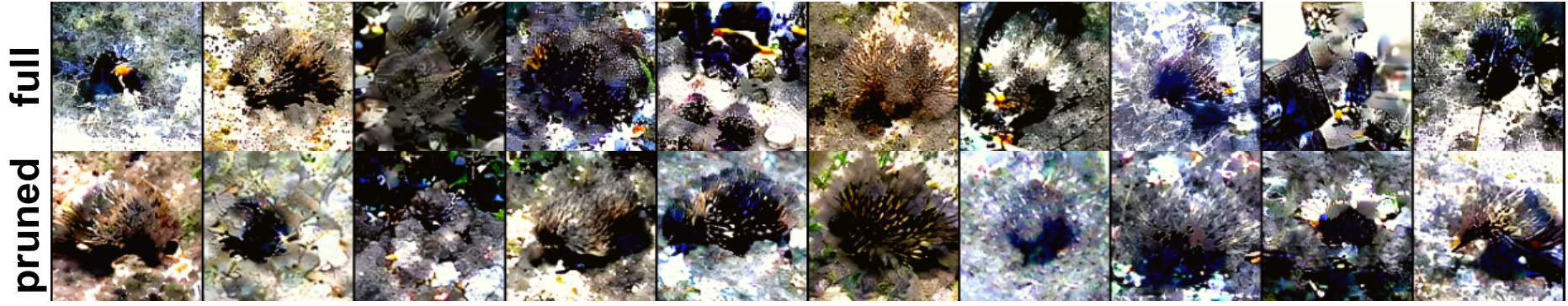}
        \caption{\label{fig:10vis} IPC=10 class images from ImageNet-B (``Echidna'', DM). Our pruning method also refines the initialization process (randomly selected image), as evidenced by the absence of humans in the bottom row.
        }
    \end{center} 
\end{figure*}

\textbf{IPC=10, 128$\times$128.}
The improved quality of distilled datasets, achieved through core-set pruning, extends to settings with IPC=10 at a resolution of 128$\times$128, as seen in \autoref{tab:10ipc}, yielding up to 3.3 percentage points improvements in ImageNet-B for LD3M with DM and 2.2 percentage points for GLaD with DC.

\textbf{Visuals for IPC=10, 128$\times$128.}
 We present the distilled images for IPC=10 in \autoref{fig:10vis}. 
 Beyond showcasing visual differences, these results highlight an additional advantage of our pruning method: typically, initial images are randomly selected from each respective class, which can introduce irrelevant or confusing features. 
 By applying pruning, we increase the likelihood of capturing clear, representative images of the actual class. For example, two images predominantly display humans are chosen from the ``Echidna'' class. Our approach pruned these two example images out as a pre-trained classifier assigned a high loss to them.

\textbf{IPC=1, 256$\times$256.}
In our 256$\times$256 experiments with IPC=1, we applied pruning to LD3M and observed a substantial improvement in cross-architecture performance across various initialization settings (ImageNet, FFHQ, and Random). 
As seen in \autoref{tab:other}, applying the LD3M models on pruned core-sets consistently outperformed their unpruned counterparts across ImageNet subsets, underscoring the effectiveness of loss-based pruning in distillation.
For instance, the pruned LD3M model initialized with ImageNet achieved an average improvement of up to +1.1 percentage points on the ImageNet-C subset and +1.5 on ImageNet-E. 
The random initialization showed similar trends, with gains of up to +1.7 percentage points on ImageNet-B. 
Thus, we can conclude that pruning also improves distillation quality for higher resolutions (\textit{i.e.}, 256$\times$256).

\begin{figure*}[!ht]
  \centering
  \includegraphics[width=.8\textwidth]{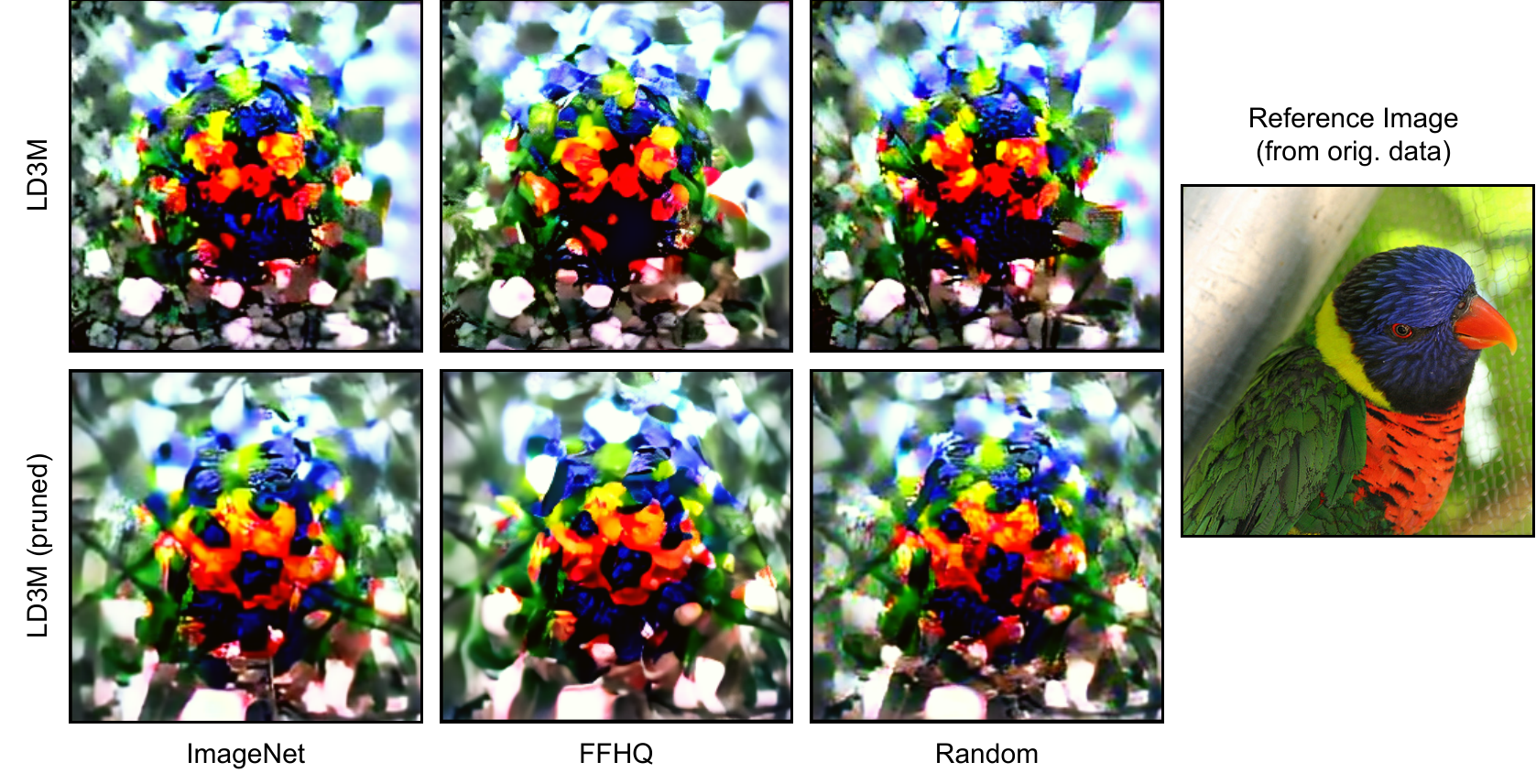}
  \caption{\label{fig:comp256}Example $256\times256$ images of a distilled class (ImageNet-B: Lorikeet) with differently initialized LD3M (pruned vs. full dataset). The various initializations, \textit{\textit{i.e.}}, which dataset was used for training the generators, are denoted at the bottom.}
\end{figure*}

\textbf{Visuals for IPC=1, 256$\times$256.} In \autoref{fig:comp256}, we present distilled images of a single class from ImageNet-B (Lorikeet) for IPC=1, 256×256 using the LD3M model with different initializations: ImageNet, FFHQ, and Random. 
The visual results illustrate that distilled images generated from pruned core-sets offer clearer and more distinct class representations by removing class-unspecific details, such as the blue background when compared to the full dataset.

\begin{table*}[!ht]
\centering
\resizebox{\textwidth}{!}{%

\begin{tabular}{llcccccc}
Distillation Space & Initialization  & All & ImNet-A   & ImNet-B   & ImNet-C   & ImNet-D   & ImNet-E   \\ \midrule
pixel space & - & 29.5{\color{gray}$\pm$3.1} & 38.3{\color{gray}$\pm$4.7} & 32.8{\color{gray}$\pm$4.1} & 27.6{\color{gray}$\pm$3.3} & 25.5{\color{gray}$\pm$1.2} & 23.5{\color{gray}$\pm$2.4} \\
\midrule

& ImageNet & 34.4{\color{gray}$\pm$2.6} & 37.4{\color{gray}$\pm$5.5} & 41.5{\color{gray}$\pm$1.2} & 35.7{\color{gray}$\pm$4.0} & 27.9{\color{gray}$\pm$1.0} & 29.3{\color{gray}$\pm$1.2}\\

& Random & 34.5{\color{gray}$\pm$1.6} & 39.3{\color{gray}$\pm$2.0} & 40.3{\color{gray}$\pm$1.7} & 35.0{\color{gray}$\pm$1.7} & 27.9{\color{gray}$\pm$1.4} & 29.8{\color{gray}$\pm$1.4} \\

\multirow{-3}{*}{GLaD} & FFHQ & 34.0{\color{gray}$\pm$2.1} & 38.3{\color{gray}$\pm$5.2} & 40.2{\color{gray}$\pm$1.1} & 34.9{\color{gray}$\pm$1.1} & 27.2{\color{gray}$\pm$0.9} & 29.4{\color{gray}$\pm$2.1} \\

\midrule\midrule

& ImageNet & 36.3{\color{gray}$\pm$1.6}& 42.1{\color{gray}$\pm$2.2} & 42.1{\color{gray}$\pm$1.5} & 35.7{\color{gray}$\pm$1.7} & 30.5{\color{gray}$\pm$1.4} & 30.9{\color{gray}$\pm$1.2} \\

& Random & 36.5{\color{gray}$\pm$1.6}& 42.0{\color{gray}$\pm$2.0} & 41.9{\color{gray}$\pm$1.7} & 37.1{\color{gray}$\pm$1.4} & 30.5{\color{gray}$\pm$1.5} & 31.1{\color{gray}$\pm$1.4} \\

\multirow{-3}{*}{LD3M} & FFHQ & 36.3{\color{gray}$\pm$1.5}& 42.0{\color{gray}$\pm$1.6} & 41.9{\color{gray}$\pm$1.6} & 36.5{\color{gray}$\pm$2.2} & 30.5{\color{gray}$\pm$0.9} & 30.6{\color{gray}$\pm$1.1} \\

\midrule

& ImageNet & 37.3{\color{gray}$\pm$1.5}  & 42.6{\color{gray}$\pm$1.9} & 43.2{\color{gray}$\pm$1.5} & 37.6{\color{gray}$\pm$1.6} & 30.8{\color{gray}$\pm$1.2} & 32.4{\color{gray}$\pm$1.2} \\
&& \color{ForestGreen}+1.0{\color{gray}-0.1} & \color{ForestGreen}+0.5{\color{gray}-0.3} & \color{ForestGreen}+1.1{\color{gray}-0.7} & \color{ForestGreen}+1.1{\color{gray}-0.6} & \color{ForestGreen}+0.3{\color{gray}+0.3} & \color{ForestGreen}+1.5{\color{gray}+0.0} \\

\textbf{LD3M} & Random & 37.4{\color{gray}$\pm$1.3}& 43.2{\color{gray}$\pm$1.5} & 43.6{\color{gray}$\pm$1.7} & 37.5{\color{gray}$\pm$1.3} & 30.5{\color{gray}$\pm$1.1} & 32.4{\color{gray}$\pm$0.8} \\
(pruned)&& \color{ForestGreen}+0.9{\color{gray}-0.3}& \color{ForestGreen}+1.2{\color{gray}-0.5} & \color{ForestGreen}+1.7{\color{gray}+0.0} & \color{ForestGreen}+0.4{\color{gray}-0.1} & 
+0.0{\color{gray}-0.4} & \color{ForestGreen}+1.3{\color{gray}-0.6} \\

& FFHQ & 37.3{\color{gray}$\pm$1.3}& 42.0{\color{gray}$\pm$1.7} & 43.6{\color{gray}$\pm$1.2} & 37.5{\color{gray}$\pm$1.7} & 31.0{\color{gray}$\pm$1.1} & 32.6{\color{gray}$\pm$1.0} \\
&& \color{ForestGreen}+1.0{\color{gray}-0.2}& 
+0.0{\color{gray}+0.1} & \color{ForestGreen}+1.7{\color{gray}-0.4} & \color{ForestGreen}+1.0{\color{gray}-0.5} & \color{ForestGreen}+0.5{\color{gray}-0.2} & \color{ForestGreen}+2.0{\color{gray}-0.1} \\

\end{tabular}
}
\caption{
256$\times$256 distilled HR images using the DC distillation algorithm and IPC=1. 
For both scenarios (LD3M and GLaD), we evaluate pre-trained generators on ImageNet \cite{deng2009imagenet}, FFHQ \cite{karras2019style}, and randomly initialized. 
}
\label{tab:other}
\end{table*}

\subsection{Pruning Analysis}
\label{sec:pa}

Prior to the experiments presented so far, we analyzed the impact of pruning settings on ImageNette, IPC=1 across the three distillation methods with LD3M: DC, DM, and MTT.
We evaluated relative dataset sizes in two ranges: fine-grained sizes from 1\% to 9\%, and broader sizes from 10\% to 100\% in increments of 10\%. 
For each relative size $r \in (0, 100)$, we apply the same relative size class-wise, maintaining a representative subset that preserves class diversity and minimizes bias.

\begin{figure}[t!]
    \begin{center}
        \includegraphics[width=\columnwidth]{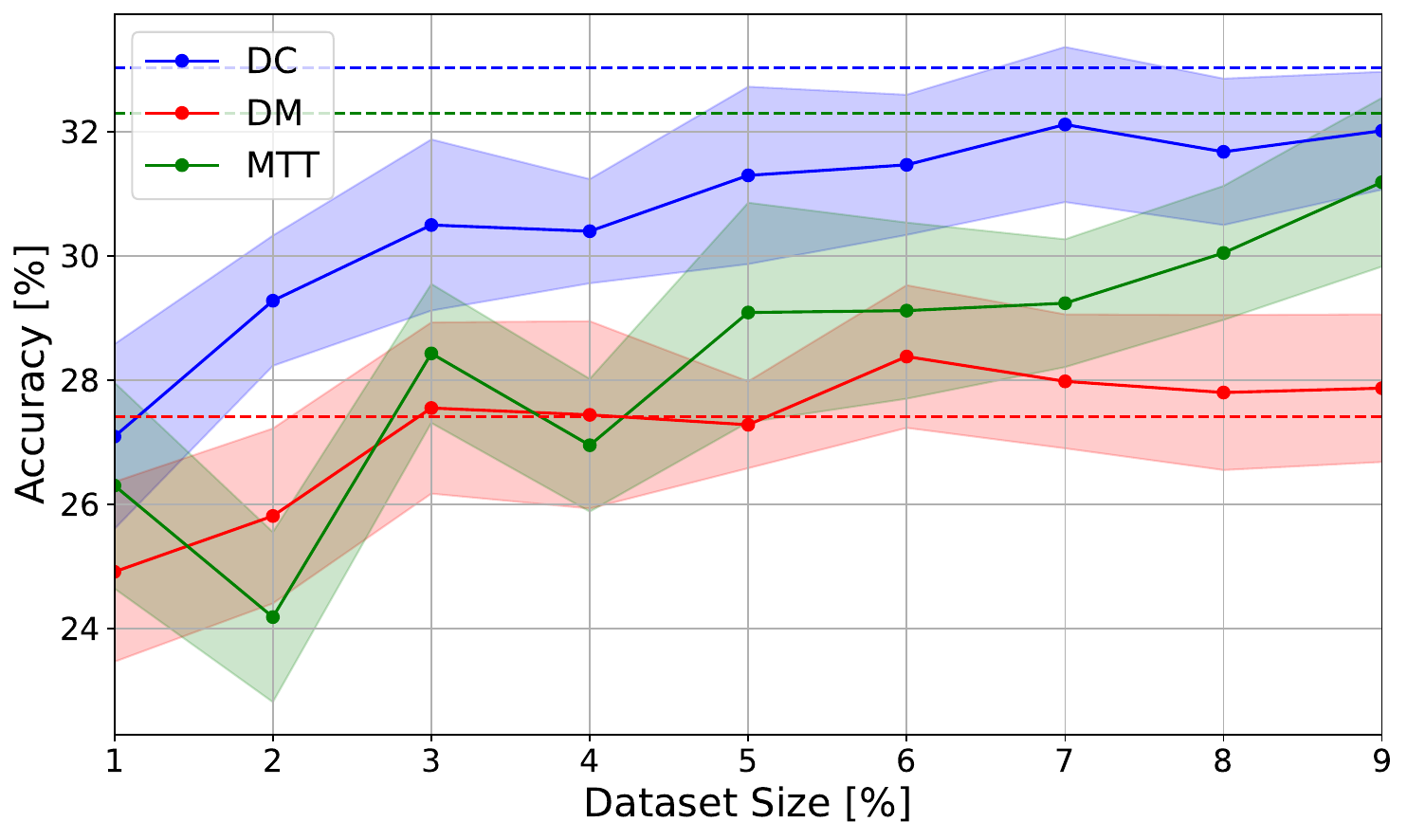}
        \caption{\label{fig:fine}Performance evaluation of DC, DM, and MTT methods on ImageNette with IPC=1 with fine-grained pruning (1-9\%) and 100\% for $\pi_r^{\text{easy}}$ (horizontal lines).
        }
    \end{center} 
\end{figure}

The results of our first investigation are shown in \autoref{fig:ascending} (more detailed Tables in the appendix). 
It shows the performance of DC, DM, and MTT methods as pruning is applied in steps from 10\% to 100\% ($\pi_r^{\text{easy}}$). 
The general trend across all methods shows that higher pruning rates (lower percentages) lead to fluctuation in the performance compared to the training on the full dataset. 
Moreover, it shows that MTT and DM can perform better than the entire dataset for various relative sizes, especially around the mid-size levels (60\%-70\%).
Surprisingly, DM seems to improve the performance as the dataset size decreases significantly (see peak at 20\%).
Thus, we took a deeper look at more fine-grained dataset sizes, specifically from 1\% to 9\%, which results are shown in \autoref{fig:fine}.

\begin{figure}[ht!]
    \begin{center}
        \includegraphics[width=\columnwidth]{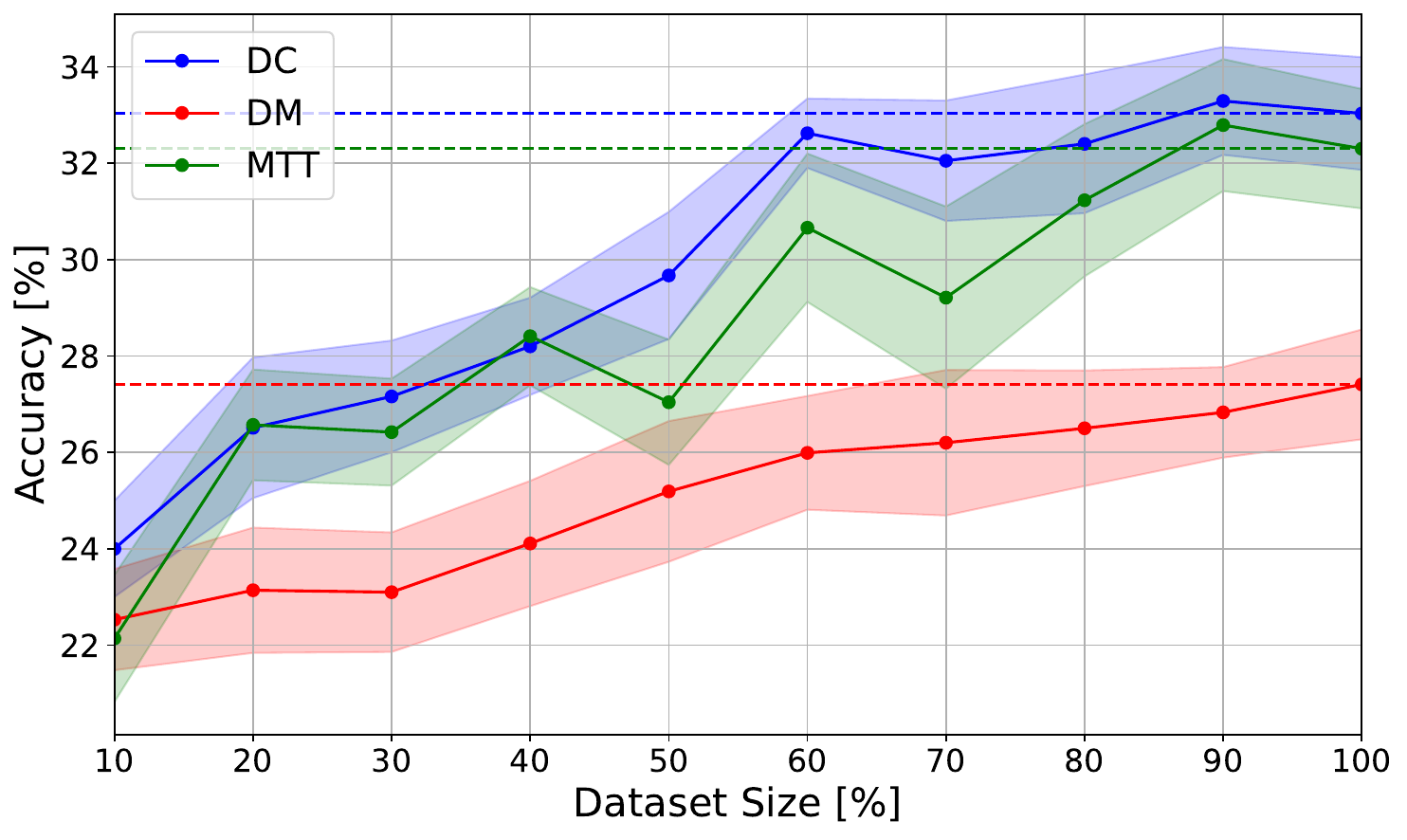}
        \caption{\label{fig:descending}Performance evaluation of DC, DM, and MTT methods on ImageNette with IPC=1. The results are shown with $\pi_r^{\text{hard}}$, horizontal lines denoting 100\%.
        }
    \end{center} 
\end{figure}

Here, all methods show a sharper decline in accuracy as the dataset size is reduced, especially for DC and MTT. 
Yet, with only 3\% of the dataset, DM can maintain the same dataset quality compared to the entire dataset. 
Lastly, \autoref{fig:descending} provides the same comparison as our first investigation but with descending pruning ($\pi_r^{\text{hard}}$).
The plot clearly shows that descending pruning consistently leads to declining dataset quality for smaller dataset sizes, highlighting the importance of easy samples, probably noisy-free samples, for dataset distillation.

\begin{figure}[t!]
    \begin{center}
        \includegraphics[width=\columnwidth]{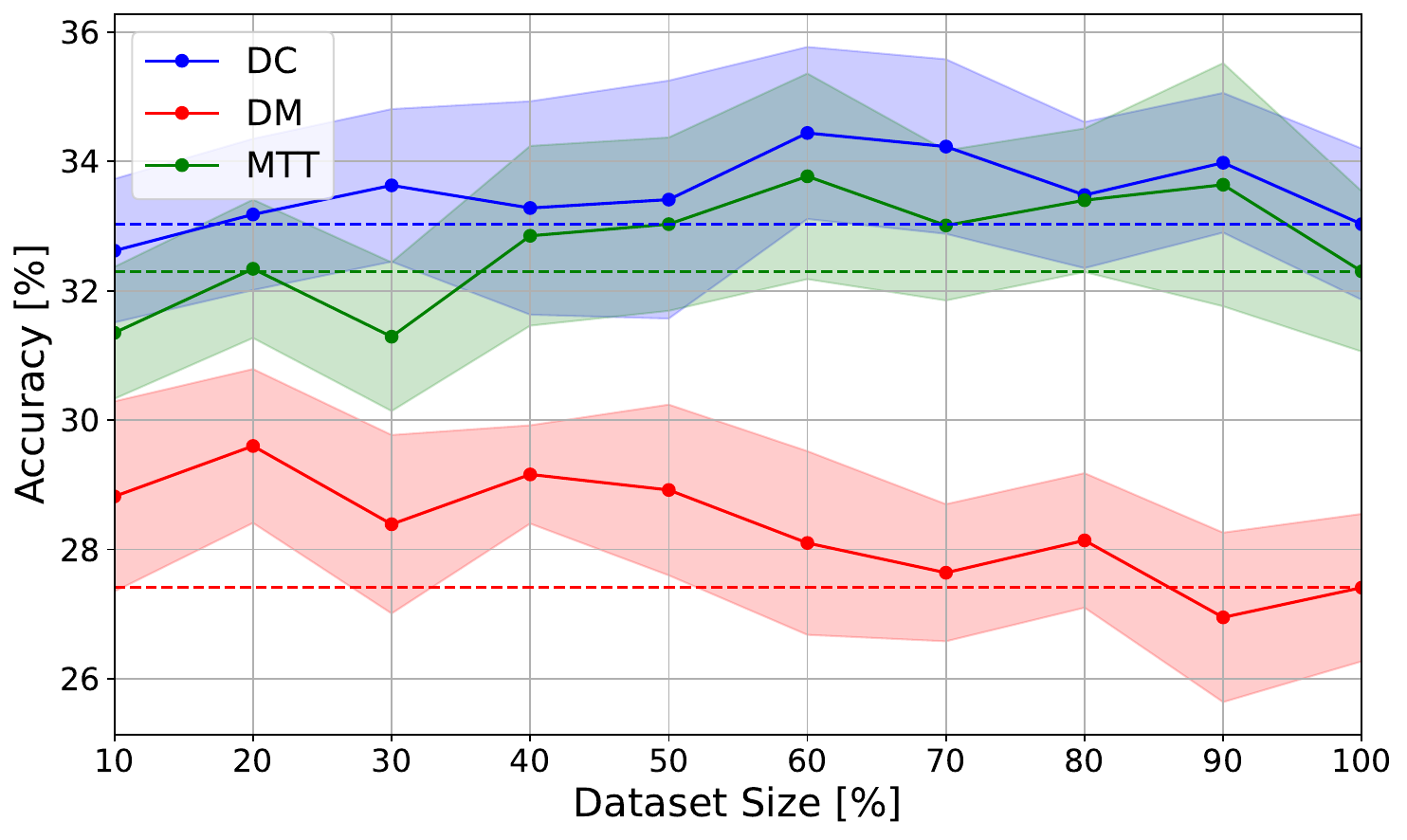}
        \caption{\label{fig:ascending}Performance evaluation of DC, DM, and MTT methods on ImageNette with IPC=1. The results are shown with $\pi_r^{\text{easy}}$, horizontal lines denoting 100\%.
        }
    \end{center} 
\end{figure}

The results highlight that while all methods experience accuracy degradation with increased pruning, they also show various pruning ranges that lead to dataset quality that maintains, and in some cases even improves, performance with reduced dataset sizes, particularly for DM. 
Conversely, descending pruning consistently reduces dataset quality, emphasizing the critical role of easier samples in achieving optimal distillation outcomes.

Overall, the optimal dataset size varies by distillation method and dataset. 
Relative sizes between 60-80\% tend to yield the best performance for DC and MTT, while DM achieves strong results with dataset sizes closer to 20\% (the values used in the experiments presented in the previous section). 
This trend suggests that each distillation method benefits from a unique balance between sample diversity and data compactness, which pruning ($\pi_r^{\text{easy}}$) helps to achieve by focusing on those contributing the most to model learning.

\section{Limitations \& Future Work}
One limitation of our approach is determining the optimal pruning factor $r$ before starting the distillation process. 
This pre-computation step requires a preliminary analysis to identify the ideal amount of data to retain in the core-set, which can add computational overhead. 
The $r$ value may vary depending on the dataset, architecture, and distillation method used, necessitating fine-tuning for each new scenario. 

Future work should develop dynamic methods for determining $r$, such as adaptive pruning strategies that adjust based on the specific characteristics of the dataset and target architecture. 
Such advancements would streamline the pruning process and make it more applicable to a broader range of datasets and architectures, further improving the generalizability of distilled datasets.

\section{Conclusion}
We introduced a novel ``Prune First, Distill After'' framework that combines loss-value-based sampling and pruning to optimize distillation quality, enhancing the performance on unseen architectures. 
We effectively address critical limitations in existing dataset distillation approaches, such as cross-architecture generalization and data redundancy, by emphasizing the importance of ``easier'' samples, \textit{i.e.}, samples with low loss values, for dataset distillation. 
Experimental results on various ImageNet subsets demonstrate that even substantial pruning (\textit{e.g.}, up to removing 80\% of the original dataset) prior to distillation maintains or, most of the time, improves distillation quality, with accuracy gains reaching up to 5.2 percentage points on various subsets. 
This robust performance across diverse architectures and pruning factors showcase the scalability and generalizability of our framework, marking a significant advancement in the field of dataset distillation.

\section*{Acknowledgements}
This work was supported by the BMBF projects SustainML (Grant 101070408), Albatross (Grant 01IW24002) and by Carl Zeiss Foundation through the Sustainable Embedded AI project (P2021-02-009).

{\small
\bibliographystyle{ieee_fullname}
\bibliography{egbib}
}

\end{document}